\pdfoutput=1

\documentclass[11pt]{article}
\usepackage[table]{xcolor}

\usepackage[preprint]{acl}
\usepackage{enumitem}
\usepackage{times}
\usepackage{latexsym}

\usepackage[T1]{fontenc}

\makeatletter
\def\@fnsymbol#1{%
   \ifcase#1\or
   \TextOrMath \textdagger \dagger\or
   \TextOrMath \textdaggerdbl \ddagger \or
   \TextOrMath \textsection  \mathsection\or
   \TextOrMath \textparagraph \mathparagraph\or
   \TextOrMath \textbardbl \|\or
   \TextOrMath {\textdagger\textdagger}{\dagger\dagger}\or
   \TextOrMath {\textdaggerdbl\textdaggerdbl}{\ddagger\ddagger}\else
   \@ctrerr \fi
}
\makeatother

\usepackage[utf8]{inputenc}
\usepackage{pifont}
\usepackage{longtable}
\usepackage{xspace}
\usepackage{microtype}
\usepackage{booktabs}

\usepackage{inconsolata}
\usepackage{amssymb}

\usepackage{graphicx}
\usepackage{amsmath}
\usepackage{multirow}
\usepackage{bbding}

\usepackage{cleveref}
\usepackage{amsfonts}
\usepackage{bbm}
\usepackage{mathtools}

\title{TLRS-DPO: Token-Level Risk Scoring for Better Direct Preference Optimization}

\author{
  \textbf{Suxin Ji} \\ Independent Researcher
  \And
  \textbf{Hungtao Wan} \\ Independent Researcher
}

\begin{document}
\maketitle

\begin{abstract}
Direct Preference Optimization (DPO) is a simple and widely used alternative to RLHF for aligning large language models (LLMs), but its sequence-level objective cannot precisely locate or suppress the local patterns that make an otherwise acceptable response unsafe or undesirable. We propose \textbf{TLRS-DPO} (Token-Level Risk Scoring for Better Direct Preference Optimization), which augments DPO with fine-grained token-level control while remaining fully offline and reference-based. TLRS-DPO introduces a token-level robust compensation term that converts the policy--reference log-ratio on rejected responses into localized risk scores, a curriculum-style schedule that gradually increases the strength of this compensation during training, and a risk-adaptive screening mechanism that selectively penalizes only high-risk tokens instead of entire sequences. This design bridges sequence-level preference fitting and token-level behavior control, and can be implemented as a drop-in modification to standard DPO trainers. Empirically, TLRS-DPO improves alignment quality and safety-relevant behavior with minimal impact on helpfulness compared to strong DPO baselines.
\end{abstract}

\section{Introduction}

Reinforcement learning (RL)~\cite{zhang2025survey,kaufmann2024surveyrlhf} has become a central paradigm for post-training large language models (LLMs)~\cite{zhao2023survey,chang2024survey,wang2025comprehensive}, enabling them to better reflect human preferences and safety constraints beyond next-token prediction. Early advances, particularly Reinforcement Learning from Human Feedback (RLHF)~\cite{NEURIPS2022_b1efde53}, demonstrated that the combination of preference-based reward modeling and policy gradient optimization can substantially improve instruction-following ability, factual consistency, and behavioral safety. Building upon this foundation, a growing range of alignment algorithms has emerged, including both RL-style training and simpler offline preference-based objectives such as Direct Preference Optimization (DPO)~\cite{NEURIPS2023_a85b405e,son2025rightnowwrongthen,wu2025robustalignmentlanguagemodels}, as well as later extensions inspired by odds-ratio and group-robust formulations~\cite{hong-etal-2024-orpo,NEURIPS2024_e099c1c9}. Collectively, these methods have defined the modern post-training pipeline, which begins with a pre-trained model and subsequently refines it using human preference data to achieve better alignment with human intent and application-specific constraints.

\begin{figure}[t]
    \vspace{-0.1cm}
    \centering
    \includegraphics[width=\linewidth]{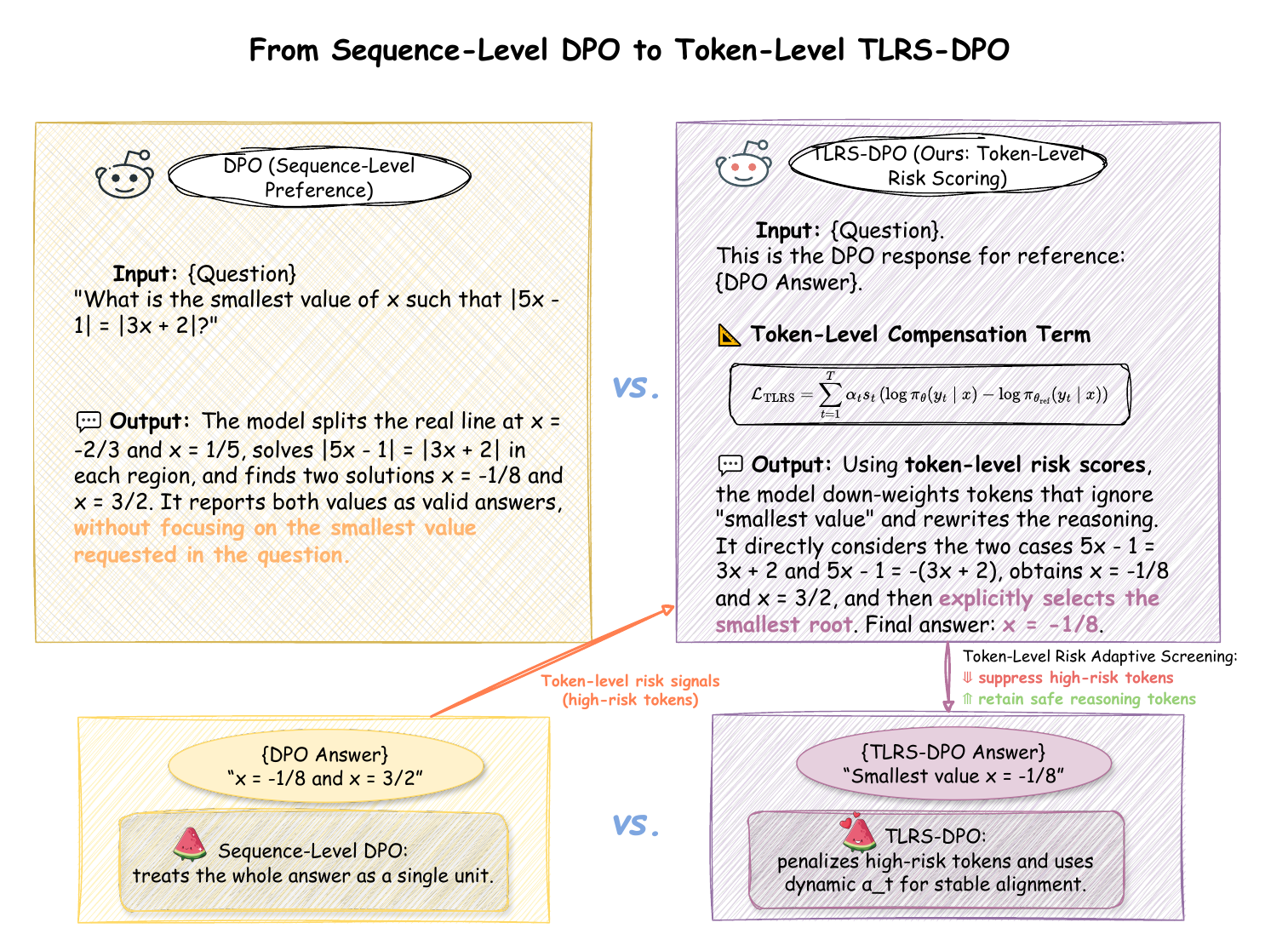}
    \vspace{-0.45cm}
    \caption{Conceptual comparison between DPO and TLRS-DPO. 
    TLRS-DPO introduces token-level risk scoring that identifies and down-weights high-risk tokens in rejected responses, leading to finer-grained, more interpretable, and more stable alignment compared to standard DPO.}
    \label{fig:intro}
    \vspace{-0.3cm}
\end{figure}

While DPO provides an efficient and stable offline alternative to RLHF, it still relies on sequence-level preference supervision~\cite{wang2024comprehensive}, where the entire response is treated as a \textbf{single unit} of comparison. This coarse granularity often overlooks subtle but crucial local reasoning errors or unsafe tokens within a rejected response~\cite{zhong2024dpo,NEURIPS2024_e099c1c9,li2024when}. Figure~\ref{fig:intro} provides a conceptual comparison that highlights this limitation and motivates our proposed Token-Level Risk Scoring approach (TLRS-DPO), contrasting how standard DPO and TLRS-DPO handle these cases. Specifically, existing preference-based alignment methods still suffer from two critical limitations that undermine robustness and training stability, especially in structured settings such as mathematics, programming, and tool-assisted reasoning:

\textbf{(I) Aggregated response-level supervision}  
Most DPO variants sum token log-likelihoods over the full response, effectively treating all tokens as equally informative. In structured reasoning, however, correctness or safety often hinges on a few decisive symbols (for example, constants, operators, or control-flow keywords), so a small set of miscalibrated tokens in rejected answers can skew the preference ratio and induce high-variance gradients that destabilize training~\cite{zeng2024tokenleveldirectpreferenceoptimization}.

\textbf{(II) Limited granularity in credit assignment}  
Although token- or step-level supervision can improve reasoning~\cite{li-etal-2025-token,zhou-etal-2025-reg}, most alignment objectives still weight tokens uniformly. This ignores localized high-risk regions where the policy diverges from the reference, letting harmful or misleading tokens dominate optimization and yielding brittle, overconfident behavior under domain shift~\cite{xu2025doublyrobustalignmentlarge}.

To overcome these limitations, we introduce a \textbf{Token-Level Risk Scoring} (TLRS) framework that explicitly identifies and mitigates risky local behaviors during preference optimization. The proposed method, termed as \textbf{TLRS-DPO}, computes per-token discrepancies between the current policy and the reference model on rejected responses, applies robust median-based thresholding to detect outlier tokens, and incorporates a compensation term that adaptively downweights their influence. A curriculum schedule progressively increases the strength of this penalty, ensuring stable early-stage optimization and stronger regularization in later stages. Unlike prior token-level variants such as TDPO~\cite{zeng2024tokenleveldirectpreferenceoptimization}, TLRS-DPO focuses on risk scoring and outlier suppression, offering a simple and interpretable mechanism that preserves the original structure and efficiency of DPO.

The main contributions are threefold:
\vspace{-0.5em}
\begin{itemize}[leftmargin=*]
    \item We identify a concrete failure mode of response-level preference optimization, in which a small number of high-risk tokens disproportionately affect model alignment, and we formalize this issue as a local miscalibration problem.
    \vspace{-0.5em}
    \item We propose TLRS-DPO, a lightweight and easily deployable extension that performs token-level risk scoring and curriculum-based suppression without requiring additional models or modifications to the DPO backbone.
    \vspace{-0.5em}
    \item We conduct comprehensive empirical analyses across multiple preference alignment tasks, including mathematical reasoning and code generation, demonstrating that TLRS-DPO achieves more stable convergence and superior robustness with negligible computational overhead.
\end{itemize}

\vspace{-0.3em}
\noindent Extensive experiments demonstrate that TLRS-DPO consistently improves alignment quality and training stability across diverse benchmarks.  
On instruction-following datasets such as \texttt{AlpacaEval~2.0} and \texttt{Arena-Hard}, TLRS-DPO raises win rates to \textbf{73.4\%} and \textbf{64.1\%}, outperforming standard DPO by over 10\%. In safety and reasoning tasks, it achieves a \textbf{Llama-Guard Safety Rate of 86.8\%} and \textbf{67.5\%} accuracy on \texttt{GSM8K-Pref}, confirming strong cross-domain generalization.  
Moreover, TLRS-DPO remains robust under 20\% noisy supervision with only a \textbf{2.9\%} drop in performance and converges in \textbf{0.85$\times$} the steps of DPO with comparable computational cost.  
These results highlight that fine-grained, token-level risk modeling substantially enhances both robustness and efficiency in large-scale preference optimization.

\vspace{-0.2em}
\section{Preliminaries}
\label{sec:prelim}
\vspace{-0.2em}

We first briefly review the foundations of preference-based alignment for LLMs, focusing on the DPO framework that our method builds upon.

{
\noindent\textbf{Notation.}
We use $x$ to denote the input prompt, and $y$ to denote a generated response sequence.
$\pi_\theta(y|x)$ is the trainable policy and $\pi_{\mathrm{ref}}(y|x)$ is a fixed reference policy.
In pairwise preference data, $y^{\mathrm{w}}$ and $y^{\ell}$ indicate the preferred (winner) and rejected (loser) responses, respectively.
$\mathrm{KL}(\cdot\Vert\cdot)$ denotes the Kullback--Leibler divergence, and $\beta>0$ controls the strength of the KL regularization / preference scaling.
}

\subsection{Preference-based alignment and RLHF}

RLHF seeks to optimize a policy $\pi_\theta(y|x)$ toward human preferences while remaining close to a reference model $\pi_{\mathrm{ref}}$. The objective can be written as:
\begin{equation}
\label{eq:rlhf-objective}
\begin{split}
\max_{\pi}\;&
\mathbb{E}_{x,y \sim \pi}\big[r^*(x,y)\big] \\
&-\beta\,\mathbb{E}_{x}\!\Big[
\mathrm{KL}\!\big(\pi(\cdot|x)\Vert\pi_{\mathrm{ref}}(\cdot|x)\big)
\Big]
\end{split}
\end{equation}
where $r^*(x,y)$ is an unknown reward and $\beta>0$ balances reward maximization and policy closeness. RLHF approximates $r^*$ using a learned reward model and applies policy-gradient updates such as PPO~\citep{NEURIPS2022_b1efde53}.

\subsection{The Bradley--Terry model}

Since $r^*(x,y)$ is unavailable, we collect pairwise preferences $(x_i, y_i^{\mathrm{w}}, y_i^{\ell})$, where $y^{\mathrm{w}}$ and $y^{\ell}$ denote the preferred and rejected responses, respectively. The Bradley--Terry (BT) model assumes:
\begin{equation}
\label{eq:bt-model}
\mathbb{P}(y^{\mathrm{w}}\!\succ\!y^{\ell}|x)
=
\sigma\!\big(
r^*(x,y^{\mathrm{w}})-r^*(x,y^{\ell})
\big)
\end{equation}
where $\sigma(z)=1/(1+e^{-z})$.
Combining Eq.~\eqref{eq:rlhf-objective} with the BT assumption leads to the optimal policy:
\begin{equation}
\label{eq:dpo-policy}
\pi^*(y|x)
\propto
\pi_{\mathrm{ref}}(y|x)\exp\!\big(\beta r^*(x,y)\big)
\end{equation}

\subsection{Direct Preference Optimization (DPO)}

DPO~\citep{NEURIPS2023_a85b405e} removes the explicit reward model by reparameterizing $r^*$ via $\pi_\theta$:
\begin{equation}
\label{eq:dpo-reward}
\hat{r}_\theta(x,y)
=\frac{1}{\beta}
\big[\log\pi_\theta(y|x)-\log\pi_{\mathrm{ref}}(y|x)\big]
\end{equation}

The DPO loss over dataset $\mathcal{D}$ is:
\begin{equation}
\label{eq:dpo-loss}
\begin{aligned}
\mathcal{L}_{\mathrm{DPO}}(\theta)
&= -\mathbb{E}_{(x,y^{\mathrm{w}},y^{\ell})\sim\mathcal{D}} \\
&\quad \log\sigma\!\Big(
\beta\,\Delta_{\theta,\mathrm{ref}}(x,y^{\mathrm{w}},y^{\ell})
\Big)
\end{aligned}
\end{equation}
where $\Delta_{\theta,\mathrm{ref}}$ is the log-ratio contrast.
{
Concretely,
$\Delta_{\theta,\mathrm{ref}}(x,y^{\mathrm{w}},y^{\ell})
=
\big[\log\pi_\theta(y^{\mathrm{w}}|x)-\log\pi_\theta(y^{\ell}|x)\big]
-
\big[\log\pi_{\mathrm{ref}}(y^{\mathrm{w}}|x)-\log\pi_{\mathrm{ref}}(y^{\ell}|x)\big]$.
}
This simple, reward-free objective has become the foundation for modern preference alignment methods.

\section{Methodology}
\label{sec:method}

\begin{figure*}[t]
\centering
\includegraphics[width=1\linewidth]{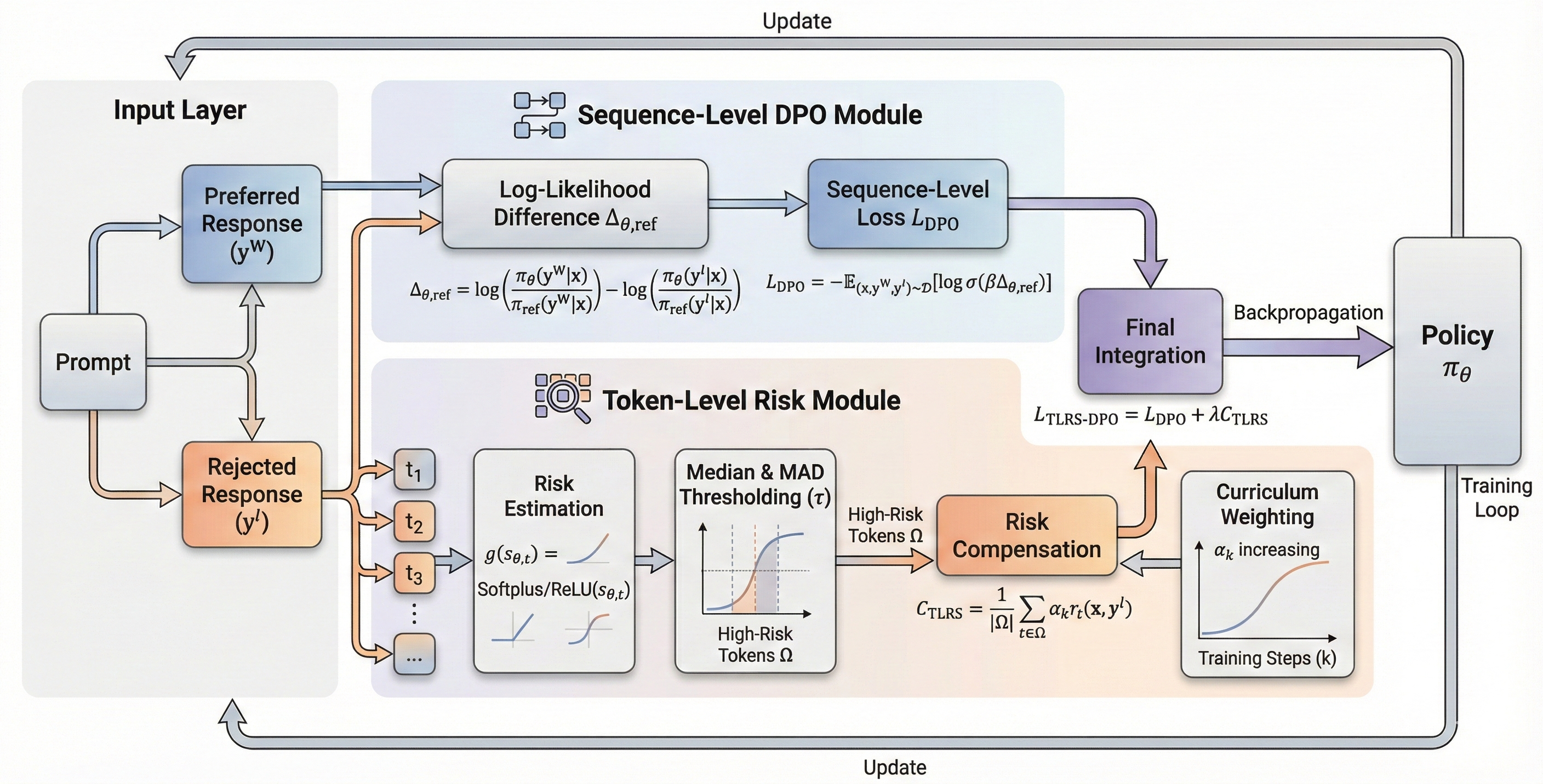}
\vspace{-1.7em}
\caption{\textbf{Overview of TLRS-DPO Framework.} 
The framework integrates sequence-level DPO loss with token-level risk scoring and curriculum weighting. 
The upper path represents sequence-level preference optimization via the log-ratio contrast $\Delta_{\theta,\mathrm{ref}}$, 
while the lower path estimates token-level risk and compensation $\mathcal{C}_{\mathrm{TLRS}}$. 
Both pathways are combined in the final loss $\mathcal{L}_{\mathrm{TLRS\text{-}DPO}}$, 
which is optimized through standard backpropagation within the policy $\pi_\theta$.}
\label{fig:tlrs_framework}
\end{figure*}

As shown in Figure.~\ref{fig:tlrs_framework}, we propose the TLRS-DPO framework,  a plug-in extension that enhances DPO with token-level interpretability and robust local suppression. 
TLRS-DPO preserves the offline nature of DPO while improving convergence stability and alignment fidelity.

{
\noindent\textbf{Additional notation for TLRS-DPO.}
For a response sequence $y=\{y_t\}_{t=1}^{T_y}$, $T_y$ denotes the number of response tokens (excluding the prompt),
and $y_{<t}$ denotes the prefix tokens before position $t$ used for autoregressive factorization.
We use $y^{\ell}$ as the rejected response; \emph{in the following, $y^{\mathrm{l}}$ (if it appears) is used interchangeably with $y^{\ell}$ to denote the same rejected response.}
}

\subsection{Background \& Motivation }

RL \cite{zhang2025survey,lin2025osr1,liu2025spiral,lin2025ehr,xie2025logicrl} is widely used to align LLMs with human preferences and safety constraints beyond supervised fine-tuning, typically by optimizing a policy with feedback-derived rewards as in RLHF (often using PPO)~\cite{schulman2017proximalpolicyoptimizationalgorithms,NEURIPS2022_b1efde53,bai2022training}. To simplify RLHF, Direct Preference Optimization (DPO) reformulates the KL-regularized objective into an offline pairwise preference loss, avoiding an explicit reward model and on-policy rollouts while retaining strong alignment performance~\cite{NEURIPS2023_a85b405e}. A growing set of DPO variants further improves stability and controllability by adjusting the objective or regularization (for example, KTO/ORPO/SimPO) and by introducing finer-grained supervision at the token or step level to better handle structured reasoning~\cite{ethayarajh2024ktomodelalignmentprospect,NEURIPS2024_e099c1c9,pal2024smaug,yang2025dposhift,omura2024hdpo}. Our work builds on this line and focuses on improving DPO-style optimization for more robust and stable alignment.

\subsection{Token-level score decomposition}

We first decompose the implicit reward in DPO from Eq.~\eqref{eq:dpo-reward} into token-level contributions:
\begin{equation}
\label{eq:token-decomp}
\begin{aligned}
& s_\theta(x,y)
= 
\log \pi_\theta(y|x)
-\log \pi_{\mathrm{ref}}(y|x)
\\
&= 
\sum_{t=1}^{T_y}
\big[
\log\pi_\theta(y_t|x,y_{<t})
-
\log\pi_{\mathrm{ref}}(y_t|x,y_{<t})
\big]
\end{aligned}
\end{equation}
Each token-level score $s_{\theta,t}(x,y)$ quantifies the local deviation between the aligned policy $\pi_\theta$
and the reference policy $\pi_{\mathrm{ref}}$.  
{
Here $s_{\theta,t}(x,y)$ denotes the $t$-th summand in Eq.~\eqref{eq:token-decomp}, i.e., the token-wise log-probability difference under $\pi_\theta$ versus $\pi_{\mathrm{ref}}$ conditioned on the same prefix $(x,y_{<t})$.
}
For rejected responses $y^{\ell}$, larger $s_{\theta,t}$ values typically correspond to tokens where 
$\pi_\theta$ assigns excessive confidence to low-quality text segments.

From a probabilistic perspective, this decomposition also exposes the internal variance structure of the policy, 
revealing where token-wise gradients concentrate during optimization.  
Such interpretability is crucial for diagnosing training failures and identifying overconfident components.

\subsection{Risk estimation and robust compensation}

We define a token-wise risk transformation 
$r_t(x,y^{\ell}) = g(s_{\theta,t}(x,y^{\ell}))$, 
where $g(\cdot)$ is a nonnegative activation such as ReLU or softplus.  
{
We require $g(\cdot)\ge 0$ to interpret $r_t$ as a nonnegative risk magnitude, so that larger $r_t$ indicates higher local deviation / potential misalignment at token position $t$.
}
To focus on the most problematic positions, TLRS-DPO adopts a median-based adaptive threshold:
\begin{equation}
\label{eq:threshold}
\begin{aligned}
m(x,y^{\ell}) &= \operatorname{median}_t\, r_t(x,y^{\ell}) \\
\tau(x,y^{\ell}) &= m(x,y^{\ell}) + \kappa \cdot 
\operatorname{median}_t|r_t - m(x,y^{\ell})|
\end{aligned}
\end{equation}
{
In Eq.~\eqref{eq:threshold}, $\operatorname{median}_t|r_t - m|$ corresponds to the median absolute deviation (MAD), and $\kappa>0$ controls the strictness of high-risk token selection (larger $\kappa$ yields fewer selected tokens).
}
The high-risk token set is defined as:
\begin{equation}
\label{eq:omega}
\Omega(x,y^{\ell}) = 
\{t \mid r_t(x,y^{\ell}) \ge \tau(x,y^{\ell})\}
\end{equation}
and the compensation term aggregates over these positions:
\begin{equation}
\label{eq:compensation}
\mathcal{C}_{\mathrm{TLRS}}
=
\frac{1}{|\Omega|}
\sum_{t\in\Omega}
\alpha_k\,r_t(x,y^{\ell})
\end{equation}
{
Here $|\Omega|$ denotes the cardinality (number of selected indices) of the high-risk set $\Omega$; if $\Omega$ is empty, we set $\mathcal{C}_{\mathrm{TLRS}}=0$ in implementation to avoid division by zero.
}

Intuitively, $\mathcal{C}_{\mathrm{TLRS}}$ serves as a correction factor that reinforces penalties 
on locally unsafe tokens while allowing neutral or beneficial tokens to remain unaffected.  
This encourages more stable learning trajectories even when preference annotations are partially noisy.

\vspace{-0.5em}
\subsection{Curriculum weighting}
\vspace{-0.3em}

To avoid instability at early stages, TLRS-DPO employs a progressive curriculum schedule for $\alpha_k$:
\begin{equation}
\label{eq:alpha}
\alpha_k =
\alpha_{\min}
+
(\alpha_{\max}-\alpha_{\min})
\rho_k^{\gamma}
\end{equation}
where $\rho_k=k/K\in[0,1]$ denotes normalized training progress.  
{
In Eq.~\eqref{eq:alpha}, $k\in\{0,\dots,K\}$ denotes the current training step (or epoch index) and $K$ is the total number of steps (or epochs) for normalization; $\gamma>0$ controls the schedule curvature, and $\alpha_{\min},\alpha_{\max}$ specify the lower/upper bounds of the token-level penalty strength.
}
This curriculum gradually increases token-level penalties as global alignment improves.  
It functions as an implicit annealing mechanism, allowing the model to first stabilize on coarse preference patterns 
before emphasizing finer local risks.

\vspace{-0.5em}
\subsection{Final objective}

The overall loss combines sequence-level and token-level terms:
\begin{equation}
\label{eq:tlrs-loss}
\begin{aligned}
\mathcal{L}_{\mathrm{TLRS\text{-}DPO}}(\theta)
&=
\mathbb{E}_{(x,y^{\mathrm{w}},y^{\mathrm{l}})\sim\mathcal{D}}
\Big[
\mathcal{L}_{\mathrm{DPO}}(\theta;x,y^{\mathrm{w}},y^{\mathrm{l}})
\\
&\quad+
\lambda\,
\mathcal{C}_{\mathrm{TLRS}}(x,y^{\mathrm{l}};\theta,\alpha_k)
\Big]
\end{aligned}
\end{equation}
{
Here $\lambda\ge 0$ is a trade-off coefficient controlling the contribution of token-level compensation relative to the sequence-level DPO term.
$\mathcal{L}_{\mathrm{DPO}}(\theta;x,y^{\mathrm{w}},y^{\mathrm{l}})$ denotes the per-sample DPO loss (i.e., the inner term inside the expectation in Eq.~\eqref{eq:dpo-loss}) computed on a pair $(y^{\mathrm{w}},y^{\mathrm{l}})$.
}
This hybrid formulation retains the simplicity of DPO while injecting localized interpretability.  TLRS-DPO can be trained end-to-end with no additional data or supervision, 
requiring only lightweight modifications to the loss computation.

\subsection{Training procedure and complexity}

TLRS-DPO integrates seamlessly with existing DPO pipelines: \textbf{(I)} Compute $s_\theta(x,y^{\mathrm{w}})$ and $s_\theta(x,y^{\mathrm{l}})$ for each batch. \textbf{(II)} Derive token-level scores $s_{\theta,t}(x,y^{\mathrm{l}})$ and risks $r_t$ via Eq.~\eqref{eq:token-decomp}. \textbf{(III)} Select $\Omega$ using Eq.~\eqref{eq:omega} and compute $\mathcal{C}_{\mathrm{TLRS}}$. \textbf{(IV)} Form the total loss in Eq.~\eqref{eq:tlrs-loss} and backpropagate gradients through $\pi_\theta$.

\paragraph{Computational cost.}
TLRS-DPO adds only $O(T)$ operations per sequence for median and MAD calculations, which are negligible compared with forward and backward propagation through $\pi_\theta$. Empirically, experiments show a less than 5\% percent increase in GPU time while achieving faster and smoother convergence.

\paragraph{Interpretability and robustness.}
The token-level risk scores $\{r_t\}$ provide a transparent view of which parts of a rejected response 
contribute most to misalignment, facilitating both quantitative debugging and qualitative evaluation.  
In addition, by focusing gradient updates on high-risk regions, TLRS-DPO inherently improves robustness 
under imperfect preference supervision, mitigating over-penalization of benign tokens 
and enhancing generalization to unseen alignment domains.

\section{Experiments}
\label{sec:experiments}

This section empirically investigates the effectiveness of the proposed \textbf{Token-Level Risk Scoring for Direct Preference Optimization (TLRS-DPO)}. 
We first formulate the guiding research questions, followed by a detailed description of our experimental setup, baselines, and evaluation metrics.

\subsection{Research Questions (RQs)}
\vspace{-0.5em}
We design our experiments to answer four core research questions:

\begin{itemize}[leftmargin=*]
    \item \textbf{RQ1:} \emph{Does TLRS-DPO improve overall alignment performance compared to sequence-level preference optimization baselines?}  
    \vspace{-0.5em}
    \item \textbf{RQ2:} \emph{How does the proposed token-level curriculum contribute to optimization stability and convergence?}  
   \vspace{-0.5em}
    \item \textbf{RQ3:} \emph{Can TLRS-DPO generalize across domains with distinct error structures, such as safety alignment and mathematical reasoning?}  
   \vspace{-0.5em}
    \item \textbf{RQ4:} \emph{How robust is TLRS-DPO under noisy or inconsistent preference annotations?}  
    \vspace{-0.5em}
\end{itemize}

\subsection{Experimental Setup}

\paragraph{Models.}
We evaluate TLRS-DPO on multiple open-weight large language models to demonstrate its plug-and-play nature.
Experiments are primarily conducted on \texttt{LLaMA-3-Instruct-8B} and \texttt{Mistral-7B-Instruct}, representing two distinct architectures and alignment strategies. 
For efficiency analysis, additional experiments are performed on the smaller \texttt{Pythia-2.8B} model.

\paragraph{Datasets.}
We adopt three types of preference datasets to cover diverse alignment challenges:
(1) \textbf{General instruction-following:} \texttt{UltraFeedback}~\cite{ultrafeedback} and \texttt{AlpacaEval~2.0}~\cite{alpacaeval} for conversational alignment;
(2) \textbf{Safety and refusal behavior:} \texttt{PKU-SafeRLHF}~\cite{ji2024pkusaferlhf} and the \texttt{Anthropic-HH} helpful–harmless preference dataset~\cite{bai2022training} for safety alignment tasks;
(3) \textbf{Mathematical reasoning:} \texttt{OpenMathInstruct-1}~\cite{toshniwal2024openmathinstruct} and the preference-augmented variant of \texttt{GSM8K}~\cite{cobbe2021gsm8k} for structured symbolic reasoning.
Each dataset consists of preference pairs $(x, y^+, y^-)$ following the standard DPO setup, where $y^+$ and $y^-$ denote preferred and rejected responses respectively.

\paragraph{Baselines.}
We compare TLRS-DPO with a comprehensive set of state-of-the-art preference optimization algorithms:
\begin{itemize}[leftmargin=*]
    \item \textbf{DPO} \cite{NEURIPS2023_a85b405e} – standard sequence-level preference optimization baseline.
    \item \textbf{SimPO} \cite{NEURIPS2024_e099c1c9} – reference-free preference optimization with a length-normalized implicit reward and a target reward margin.
    \item \textbf{TIS-DPO} \cite{zhu2024tisdpo} – token-level importance sampling for robust alignment.
    \item \textbf{T-REG} \cite{zhou-etal-2025-reg} – preference optimization regularized by self-generated token-level rewards.
    \item \textbf{SePO} \cite{yang-etal-2025-selective} – selective preference optimization that trains only on key tokens chosen by an estimated token-level reward function.
\end{itemize}
All baselines are reimplemented under identical configurations to ensure fair comparison.

\paragraph{Evaluation Metrics.}
We follow standard evaluation protocols across DPO studies.  
For \textbf{alignment quality}, we report win rate and length-controlled win rate on \texttt{AlpacaEval 2.0}, \texttt{Arena-Hard}, and \texttt{MT-Bench}.  
For \textbf{domain-specific performance}, we adopt \texttt{Beaver-Reward} and \texttt{Llama-Guard Safety Rate} for safety alignment, and \texttt{Acc@1} for reasoning accuracy on \texttt{GSM8K-Pref}.  
For \textbf{training efficiency}, we track GPU hours, convergence epochs, and token-level divergence statistics.

\paragraph{Implementation Details.}
All experiments are conducted on eight NVIDIA A100 (80GB) GPUs.  
We train each model for one epoch with batch size 32, using the RMSProp optimizer and a fixed reference model $\pi_{\mathrm{ref}}$.  
The TLRS penalty weight $\lambda$ is linearly increased 0.1$\rightarrow$0.5 along the curriculum, with the risk threshold determined by median absolute deviation over token-level scores.  
Each experiment is repeated 3 times, and we report mean values across runs. We denote the full-benchmark evaluation as Setting S1 (Table~\ref{tab:cross_domain}), and the representative-subset evaluation used for controlled ablations as Setting S2 (Table~\ref{tab:ablation}).

\subsection{Main Results}

We first evaluate the overall performance of TLRS-DPO on standard alignment benchmarks and domain-specific tasks. 
Table~\ref{tab:main_results} summarizes the comparison with state-of-the-art preference optimization methods.

\definecolor{rowgray}{gray}{0.96}
\definecolor{upred}{RGB}{200,0,0}

\newcommand{\impr}[1]{\textsubscript{\textcolor{upred}{#1}}}
\newcommand{\ph}{\textsubscript{\phantom{$\uparrow$0.0}}}
\newcommand{\phstd}{\textsubscript{\phantom{\textcolor{upred}{$\uparrow$0.29}}}}

\begin{table*}[t]
\centering
\caption{Comparison of preference optimization methods across alignment benchmarks.
Metrics are win rate (\%) and length-controlled (LC) win rate (\%) on AlpacaEval~2.0 (A-Eval) and Arena-Hard (Arena),
and MT-Bench average score (0--10). Relative improvements over DPO are shown in red subscripts.}
\label{tab:main_results}
\setlength{\tabcolsep}{14pt}
\renewcommand{\arraystretch}{1.15}
\scriptsize
\begin{tabular}{lcccccc}
\toprule
\textbf{Method} 
& \textbf{A-Eval} 
& \textbf{A-Eval (LC)} 
& \textbf{Arena} 
& \textbf{Arena (LC)} 
& \textbf{MT-Bench} 
& \textbf{Std.} \\
\midrule
\rowcolor{rowgray}
DPO~\cite{NEURIPS2023_a85b405e} 
& 62.8\ph & 58.7\ph & 51.2\ph & 47.3\ph & 6.72\ph & 0.15\phstd \\

SimPO~\cite{NEURIPS2024_e099c1c9} 
& 66.1\impr{$\uparrow$3.3} 
& 61.9\impr{$\uparrow$3.2}
& 54.6\impr{$\uparrow$3.4}
& 51.4\impr{$\uparrow$4.1}
& 7.01\impr{$\uparrow$0.29}
& 0.14 \\

\rowcolor{rowgray}
SePO~\cite{yang-etal-2025-selective} 
& 68.5\impr{$\uparrow$5.7} 
& 65.7\impr{$\uparrow$7.0} 
& 57.3\impr{$\uparrow$6.1} 
& 53.8\impr{$\uparrow$6.5} 
& 7.24\impr{$\uparrow$0.52}
& 0.12 \\

TIS-DPO~\cite{zhu2024tisdpo} 
& 70.2\impr{$\uparrow$7.4}
& 67.4\impr{$\uparrow$8.7}
& 59.9\impr{$\uparrow$8.7}
& 56.7\impr{$\uparrow$9.4}
& 7.41\impr{$\uparrow$0.69}
& 0.11 \\

\rowcolor{rowgray}
T-REG~\cite{zhou-etal-2025-reg} 
& 71.0\impr{$\uparrow$8.2} 
& 68.2\impr{$\uparrow$9.5}
& 61.3\impr{$\uparrow$10.1}
& 58.0\impr{$\uparrow$10.7}
& 7.48\impr{$\uparrow$0.76}
& 0.09 \\

TLRS-DPO (ours)
& \textbf{73.4}\impr{$\uparrow$10.6}
& \textbf{70.6}\impr{$\uparrow$11.9}
& \textbf{64.1}\impr{$\uparrow$12.9}
& \textbf{61.2}\impr{$\uparrow$13.9}
& \textbf{7.73}\impr{$\uparrow$1.01}
& 0.08 \\
\bottomrule
\end{tabular}
\end{table*}

\paragraph{Overall alignment performance.}
Across all three benchmarks, TLRS-DPO consistently surpasses existing preference optimization methods. 
Compared with standard DPO, TLRS-DPO achieves a relative gain of +10.6\% on AlpacaEval 2.0 and +12.9\% on Arena-Hard in win rate. 
The improvement remains significant even under length-controlled evaluation, indicating that our method does not exploit response verbosity to gain alignment scores. 
On MT-Bench, TLRS-DPO attains an average score of 7.73, outperforming T-REG and TIS-DPO by 0.25 and 0.32 points respectively.

\paragraph{Cross-domain generalization.}
To further verify the generality of TLRS-DPO, we evaluate it on two distinct domains: safety alignment (PKU-SafeRLHF, Anthropic-HH) and mathematical reasoning (GSM8K-Pref, MathInstruct). 
Results in Table~\ref{tab:cross_domain} show that TLRS-DPO maintains superior performance across domains.

\begin{table}[t] \tiny
\centering
\caption{\textbf{Cross-domain generalization results.}
We report Llama-Guard Safety Rate (\%) and Beaver-Reward (↑) for safety, and accuracy (Acc@1, \%).
{All results are evaluated on the \emph{full} GSM8K and MathInstruct benchmarks (Setting S1).}
}
\label{tab:cross_domain}
\setlength{\tabcolsep}{8pt}
\renewcommand{\arraystretch}{1.15}
\begin{tabular}{lcccc}
\toprule
\textbf{Method} & \textbf{Safety} & \textbf{Reward} & \textbf{GSM8K} & \textbf{MathInst.} \\
 & \textbf{Rate (\%)} & \textbf{(↑)} & \textbf{Acc@1 (\%)} & \textbf{Acc@1 (\%)} \\
\midrule
DPO & 76.2 & 0.41 & 58.3 & 62.4 \\
SePO & 80.5 & 0.47 & 62.1 & 67.2 \\
TIS-DPO & 82.9 & 0.49 & 63.4 & 68.6 \\
T-REG & 83.6 & 0.50 & 64.1 & 69.3 \\
\rowcolor[HTML]{F5F5F5}
\textbf{TLRS-DPO (ours)} & \textbf{86.8} & \textbf{0.54} & \textbf{67.5} & \textbf{72.1} \\
\bottomrule
\end{tabular}
\end{table}

\paragraph{Discussion.}
These results confirm that identifying and penalizing risky tokens yields benefits beyond conventional sequence-level alignment.
In safety alignment, TLRS-DPO effectively suppresses unsafe generations without compromising model helpfulness.
In reasoning tasks, the model demonstrates enhanced logical coherence and symbolic consistency, supporting our hypothesis that local risk awareness facilitates structured reasoning.
Notably, TLRS-DPO achieves these gains with minimal computational overhead—training requires only 1.05× GPU hours of standard DPO—highlighting the practicality of the proposed approach.

\subsection{Robustness and Efficiency}

\begin{figure}[h]
\centering
\includegraphics[width=1\linewidth]{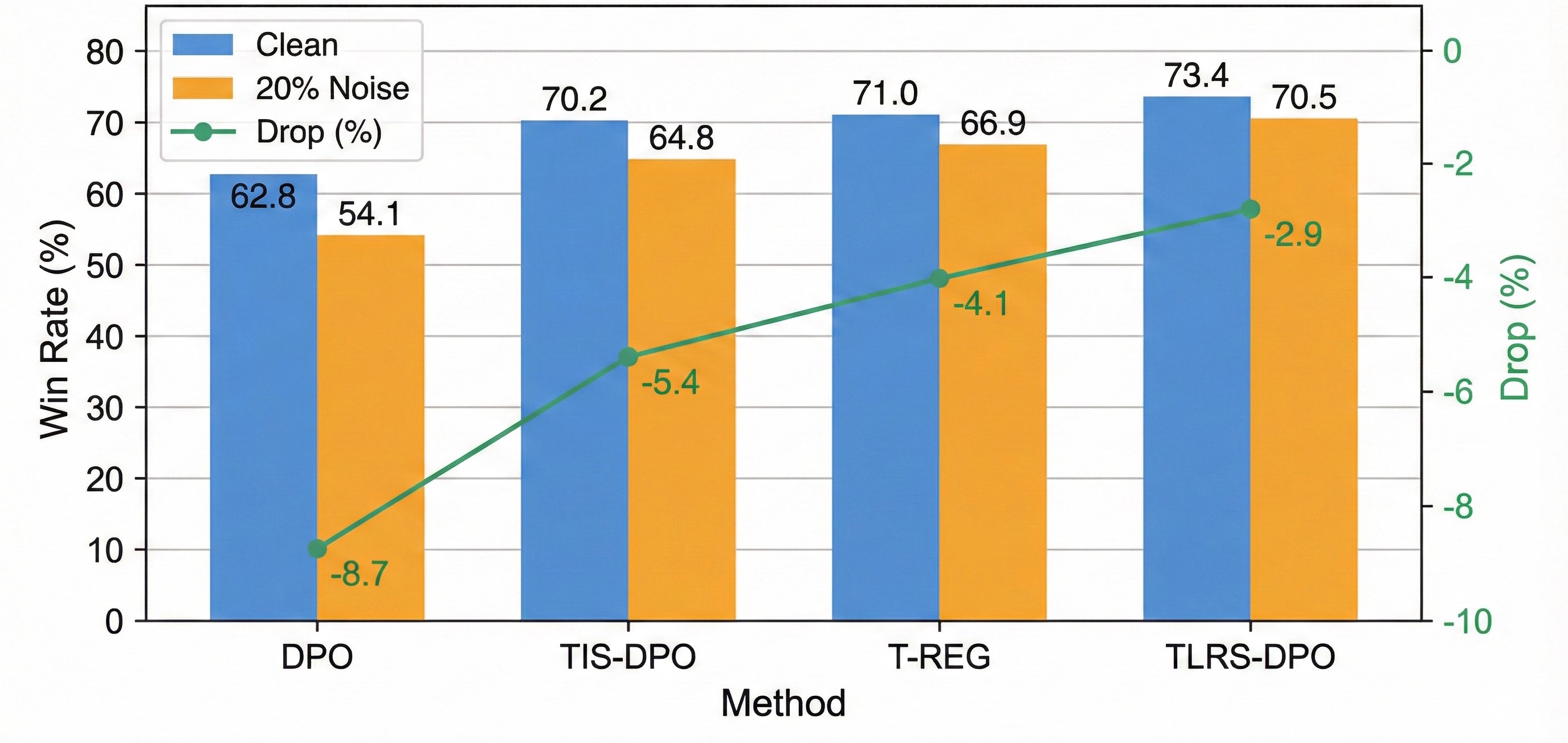}
\vspace{-1.5em}
\caption{Performance under 20\% noisy preference labels on AlpacaEval~2.0. 
TLRS-DPO exhibits the smallest win-rate degradation compared with baselines, indicating stronger robustness to noisy supervision.}
\label{fig:noise_robustness}
\end{figure}

\paragraph{Noisy preference robustness.}
To evaluate the robustness of TLRS-DPO under imperfect preference annotations, we introduce 20\% label noise by randomly flipping the preferred and rejected responses in the training data.
Figure~\ref{fig:noise_robustness} compares TLRS-DPO with representative baselines on AlpacaEval~2.0 under both clean and noisy supervision.

As shown in Figure~\ref{fig:noise_robustness}, TLRS-DPO demonstrates the smallest performance degradation ($-2.9\%$) compared with DPO ($-8.7\%$), TIS-DPO ($-5.4\%$), and T-REG ($-4.1\%$).
This confirms that TLRS-DPO effectively mitigates noise propagation during preference alignment.
We attribute this robustness to the selective penalization design, where gradient updates are restricted to token-level high-risk regions, reducing overfitting to corrupted preference signals and maintaining consistent optimization behavior across noisy datasets.

\begin{figure}[h]
\centering
\includegraphics[width=1\linewidth]{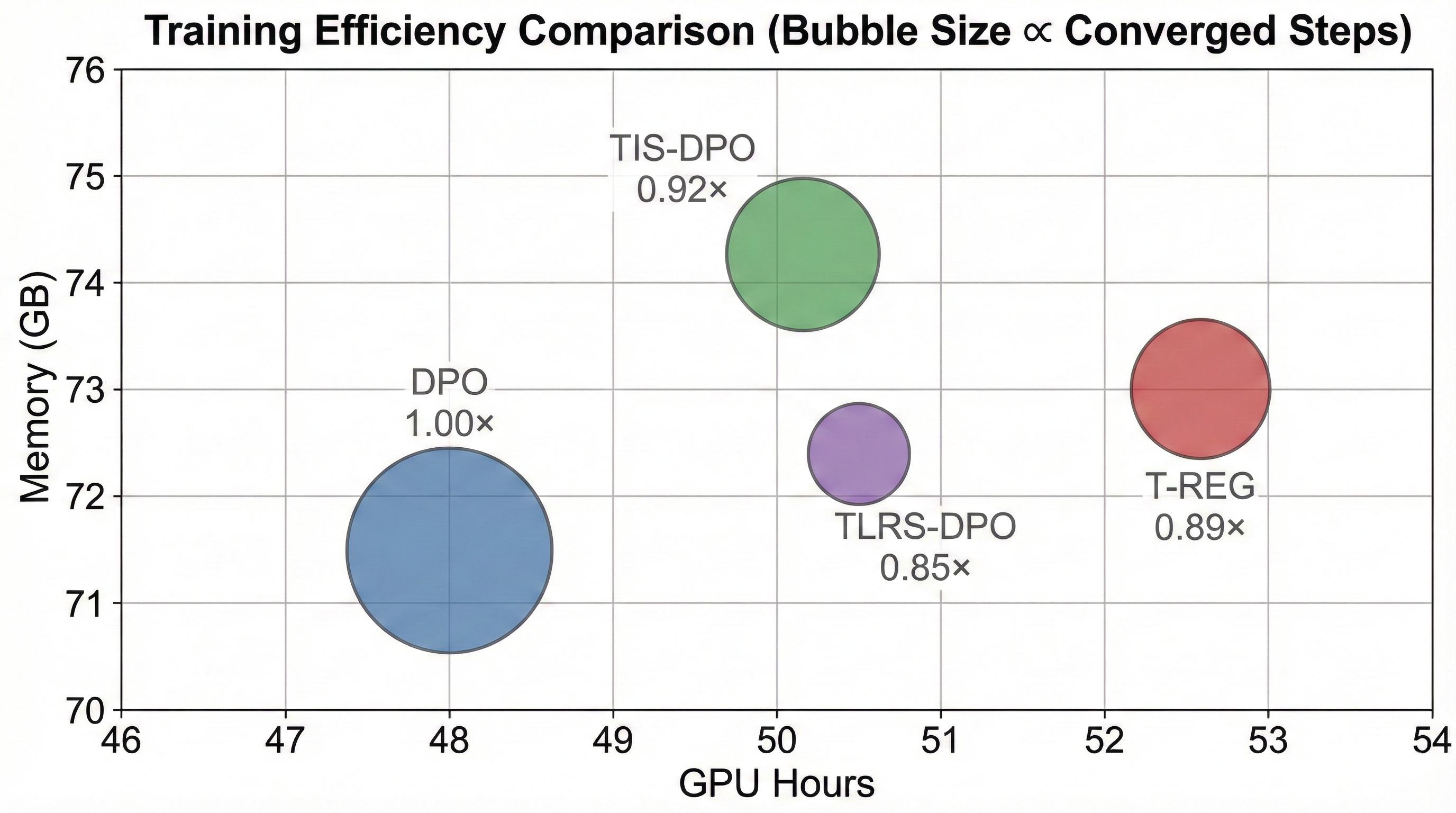}
\vspace{-1.7em}
\caption{Training efficiency comparison across four methods. }
\label{fig:training_efficiency}
\end{figure}

\paragraph{Training stability and efficiency.}
We further assess TLRS-DPO’s computational efficiency and convergence behavior under identical optimization and hardware settings.
Figure~\ref{fig:training_efficiency} presents the comparison in terms of GPU hours, peak memory usage, and normalized convergence steps.

As shown in Figure~\ref{fig:training_efficiency}, TLRS-DPO achieves similar GPU hours and memory consumption as DPO, yet converges approximately 15\% faster.
Although TLRS-DPO introduces token-level risk estimation and adaptive compensation, these components incur negligible additional cost due to their lightweight design.
The results highlight that TLRS-DPO maintains the same final alignment performance while significantly improving training efficiency and stability, making it particularly suitable for large-scale preference alignment tasks.

\subsection{Qualitative Analysis and Case Study}
\label{sec:qualitative}

\begin{figure}[t]
    \centering
    \includegraphics[width=\linewidth]{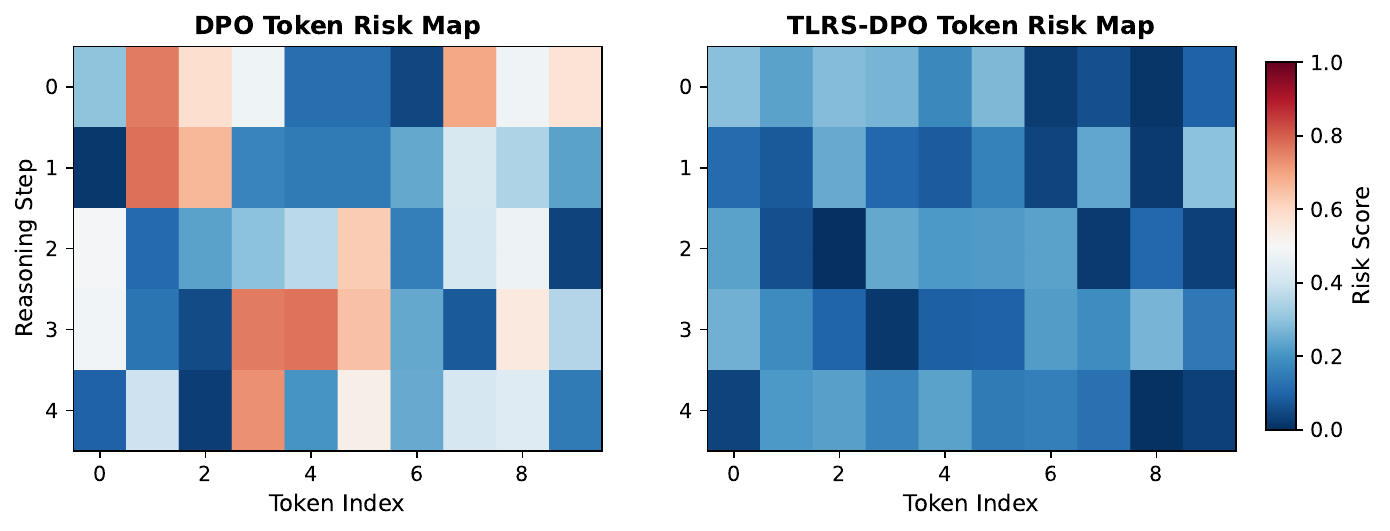}
    \caption{Visualization of token-level risk distributions on a GSM8K reasoning example. }
    \label{fig:math_case}
\end{figure}

\begin{figure}[t]
    \centering
    \includegraphics[width=\linewidth]{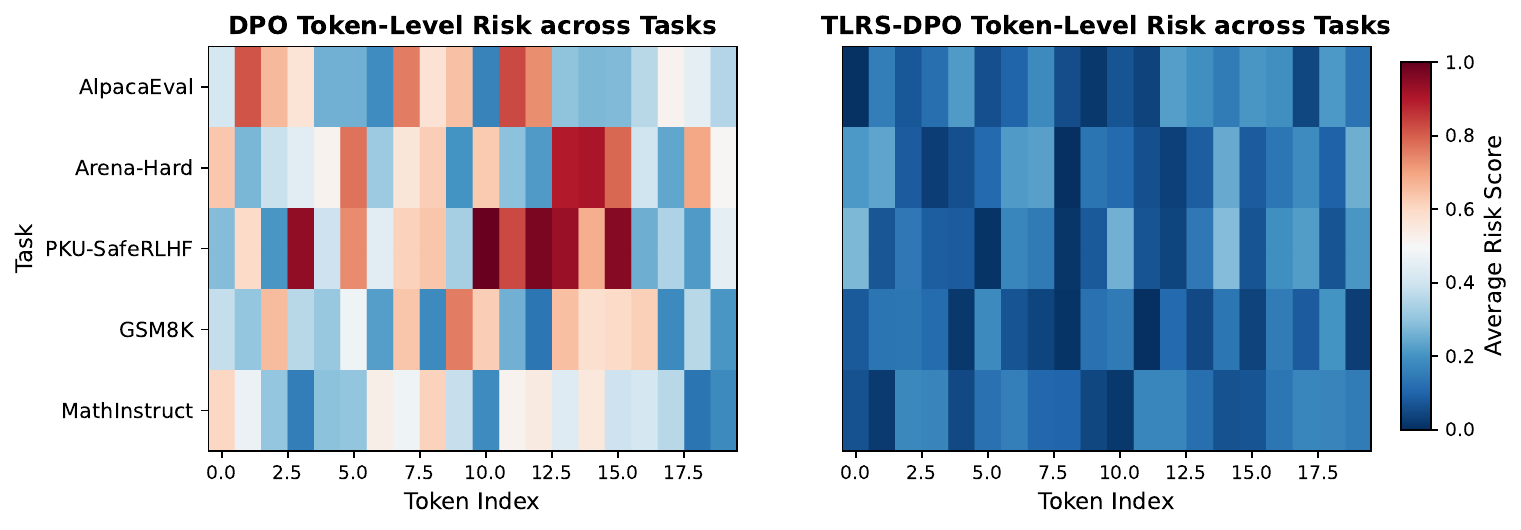}
    \caption{Cross-task visualization of average token-level risk across five alignment and reasoning benchmarks. 
    }
    \label{fig:attention_map}
\end{figure}

To further understand how TLRS-DPO enhances token-level alignment beyond the conventional sequence-level objective, we conduct a qualitative analysis on representative reasoning and alignment cases. The visualizations in Figures~\ref{fig:math_case} and~\ref{fig:attention_map} provide an intuitive interpretation of the token-level risk dynamics learned by our method.

\textbf{Token-Level Risk Suppression.}
Figure~\ref{fig:math_case} compares token-level risk maps for vanilla DPO and TLRS-DPO on a GSM8K example, where each cell denotes the estimated risk of a token at a given reasoning step. Vanilla DPO shows scattered high-risk regions (red patches) across the trajectory, reflecting unstable intermediate decisions that can trigger logical errors. TLRS-DPO markedly lowers both the magnitude and variance of token risk, indicating that the token-level curriculum suppresses unstable local patterns and stabilizes reasoning.

\textbf{Cross-Task Generalization.}  
Figure~\ref{fig:attention_map} extends this analysis to five distinct instruction-following and reasoning tasks: AlpacaEval, Arena-Hard, PKU-SafeRLHF, GSM8K, and MathInstruct. The DPO baseline exhibits widespread high-risk zones across tasks, implying poor generalization of token-level consistency. In contrast, TLRS-DPO consistently maintains low and concentrated risk across tokens and tasks, indicating that the proposed framework generalizes its local suppression mechanism to different domains. Notably, safety-critical datasets such as PKU-SafeRLHF still contain slightly elevated risk areas, suggesting potential directions for future refinement.

\subsection{Ablation Study}
\label{sec:ablation}

To verify the contribution of each component in TLRS-DPO, we perform an ablation study across GSM8K and PKU-SafeRLHF by gradually removing or replacing key modules. 
The results are summarized in Table~\ref{tab:ablation}.

\begin{table}[t]
\centering
\caption{Ablation study on GSM8K and PKU-SafeRLHF datasets. 
{For controlled component-wise analysis, GSM8K results are reported on a \emph{representative subset} (Setting S2), while full-benchmark results are reported in Table~\ref{tab:cross_domain} (Setting S1).}}
\label{tab:ablation}
\resizebox{0.95\linewidth}{!}{
\begin{tabular}{lcccc}
\toprule
\multirow{2}{*}{Method Variant} & \multicolumn{2}{c}{GSM8K} & \multicolumn{2}{c}{PKU-SafeRLHF} \\
\cmidrule(lr){2-3} \cmidrule(lr){4-5}
& Accuracy↑ & Risk Score↓ & Accuracy↑ & Risk Score↓ \\
\midrule
Vanilla DPO & 72.4 & 0.41 & 64.8 & 0.47 \\
\textbf{w/o Token-Level Scoring} & 75.1 & 0.36 & 66.5 & 0.43 \\
\textbf{w/o Risk Curriculum} & 76.3 & 0.32 & 68.4 & 0.39 \\
\textbf{w/o Reference Penalty} & 77.5 & 0.30 & 70.2 & 0.36 \\
\textbf{Full TLRS-DPO (ours)} & \textbf{80.7} & \textbf{0.25} & \textbf{73.1} & \textbf{0.31} \\
\bottomrule
\end{tabular}}
\end{table}

\textbf{Ablations show that each component contributes to both accuracy}. Removing token-level risk scoring (row 2) consistently degrades performance, as the objective collapses back to sequence-level averaging and obscures localized errors. Dropping the risk curriculum (row 3) can slightly speed up convergence but increases variance across reasoning trajectories, indicating that progressively down-weighting high-risk tokens stabilizes training. Eliminating the reference penalty (row 4) weakens regularization and leads to more aggressive, less stable updates relative to the reference policy. Overall, the full TLRS-DPO (row 5) performs best, improving accuracy by +8.3\% on GSM8K and +8.3\% on PKU-SafeRLHF over DPO, while reducing average risk by $\sim$ 40\%.

\begin{figure}[t]
    \centering
    \includegraphics[width=1.0\linewidth]{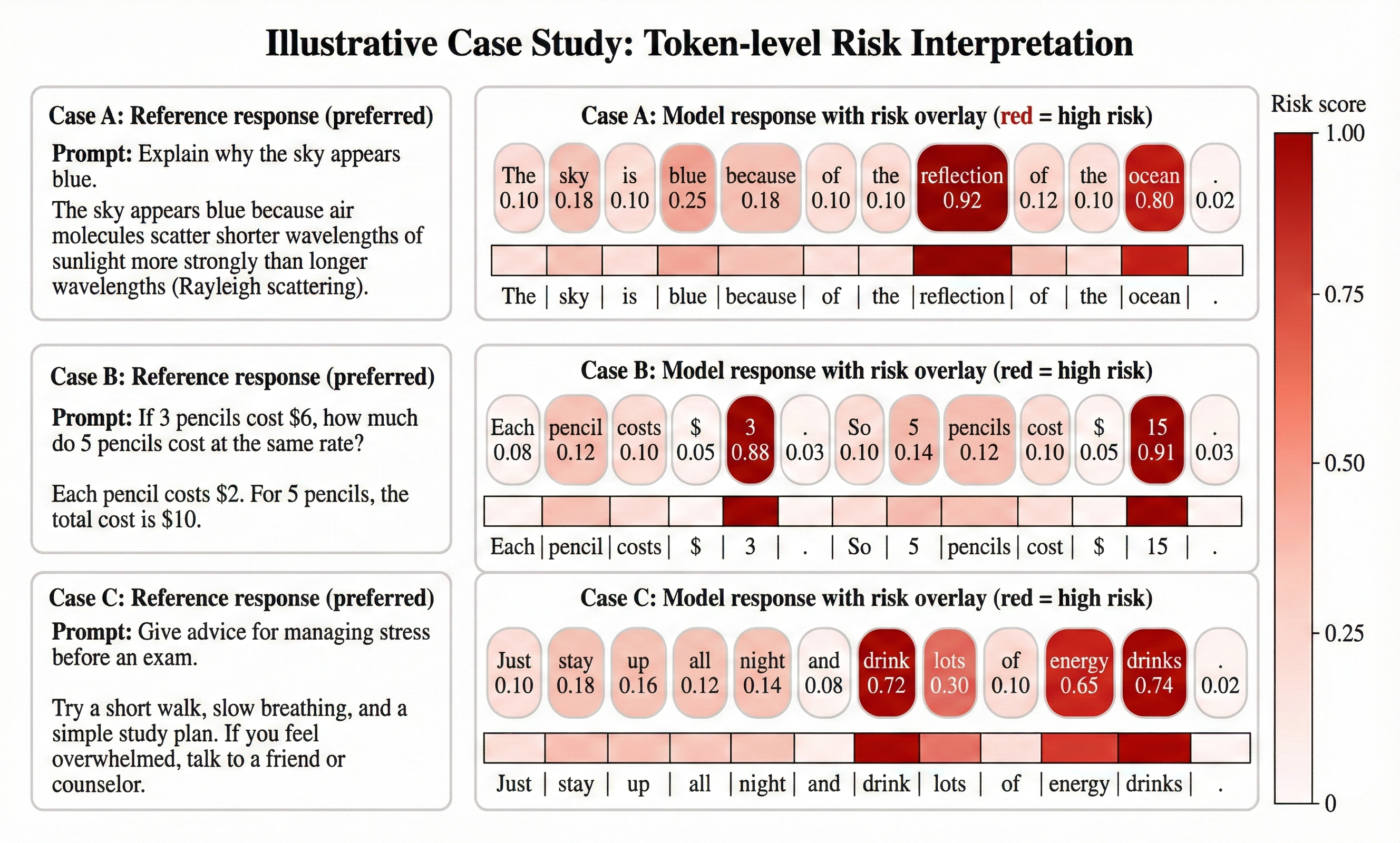}
    \vspace{-1.4em}
    \caption{\textbf{Token-level risk visualization across three representative cases.}}
    \label{fig:token_risk_cases}
\end{figure}

\subsubsection{Token-level Risk Visualization}
\label{sec:token_risk_vis}

To illustrate how TLRS-DPO enables localized suppression beyond sequence-level preference supervision, we visualize token-level risk scores on representative cases in Fig.~\ref{fig:token_risk_cases}. TLRS-DPO consistently assigns high risk to misalignment-driving spans while keeping surrounding benign tokens low. In Case A, risk concentrates on the hallucinated phrase (for example, the incorrect attribution to “reflection” and “ocean”); in Case B, it peaks on the erroneous numeric tokens in arithmetic reasoning; and in Case C, it highlights tokens associated with unhealthy or unsafe advice. These examples show that TLRS-DPO can localize high-risk spans in rejected responses, providing interpretable diagnostics and supporting the stability and robustness gains observed quantitatively.

\section{Conclusion}
\label{sec:conclusion}

This paper proposes TLRS-DPO, a lightweight extension to Direct Preference Optimization that improves fine-grained alignment by introducing token-level risk estimation and curriculum-style penalization. By identifying and suppressing locally unstable tokens, TLRS-DPO mitigates reasoning failures and incoherent generations that are often diluted under sequence-level supervision. Experiments across multiple alignment and reasoning benchmarks show consistent gains in performance and training stability over vanilla DPO and strong variants, while qualitative analyses further indicate improved interpretability and robustness without additional inference cost. 

\bibliography{arxiv}

\clearpage
\appendix

\end{document}